\documentclass[runningheads]{llncs}
\usepackage[breaklinks=true,colorlinks,bookmarks=false]{hyperref}
\usepackage{amssymb}
\usepackage{amsmath}
\usepackage{psfrag}
\usepackage{epsfig}
\usepackage{cite}
\usepackage{graphics}
\usepackage{graphicx}
\usepackage{color}
\usepackage{subfigure}
\usepackage{multirow}
\usepackage{threeparttable}
\usepackage{booktabs}
\usepackage{mmstyles}
\usepackage{bm}
\usepackage{bbm}
\usepackage[T1]{fontenc}

\usepackage{graphicx}
\begin{document}
%

\title{ImplantMamba: Long-range Sequential Modeling Mamba For Dental Implant
Position Prediction}


\titlerunning{ImplantMamba}
%
%

\author{
 Xinquan Yang\inst{1,2},
 Congmin Wang\inst{3},
 Xuguang Li\inst{4} ,
 Yulei Li\inst{3},
 Linlin Shen\inst{1,2},
 Yongqiang Deng\inst{4}
 He Meng\inst{4}
 }

\institute{School of Artifical Intelligence, Shenzhen University, Shenzhen, China \\ \and National Engineering Laboratory for Big Data System Computing Technology, Shenzhen University, China \\ \and Huangpu People's Hospital, Zhongshan City, China \and Department of Stomatology, Shenzhen University General Hospital, Shenzhen, China \\
\email{yangxinquan2021@email.szu.edu.cn, 516746581@qq.com, llshen@szu.edu.cn}
}

\maketitle              
\begin{abstract}
In the design of surgical guides for implant placement, determining the precise implant position is a critical step. However, the implant region itself is often characterized by a lack of distinctive texture in medical images. Consequently, artificial intelligence (AI) models must infer the correct implant position and angulation (slope) primarily by analyzing the texture of the surrounding teeth, which poses a significant challenge.
To address this, we propose ImplantMamba, a network architecture designed for long-range sequential modeling to integrate texture information from adjacent teeth. Our approach explicitly couples the regression of the implant position with its slope. The core of ImplantMamba is a hybrid encoder that combines Convolutional Neural Networks (CNNs) with Mamba layers. This design enables the network to hierarchically extract local anatomical features through CNNs while simultaneously modeling global contextual dependencies across the entire scan volume via Mamba's selective scan operations, leading to a more comprehensive understanding of the implant site.
Furthermore, we introduce a Slope-Coupled Prediction Branch (SCP). This branch is designed to connect the prediction of implant position with the slope, ensuring internal consistency and anatomical plausibility by thereby enforcing a coherent relationship between the predicted implant location and its angulation.
Extensive experiments on a large-scale dental implant dataset demonstrate that the proposed ImplantMamba achieves superior performance compared to existing methods.

\end{abstract}

\section{Introduction}
Tooth loss and dental fractures represent widespread global oral health challenges, making dental implantation a critical restorative procedure~\cite{elani2018trends,nazir2020global}. In modern practice, surgical guides are frequently employed to ensure precise implant placement as planned. However, the current design of these guides often relies on manual and operator-dependent analysis of Cone-Beam Computed Tomography (CBCT) data, which can introduce variability and inefficiency~\cite{kernen2020review}. 
Deep learning (DL) presents a transformative opportunity to automate implant planning by determining the optimal three-dimensional position (including location, angulation, and depth) of an implant based on the patient's anatomical structures~\cite{liu2021transfer,yang2023two}. Such automation can significantly enhance the accuracy and efficiency of guided surgery, while also facilitating patient-specific guide design, virtual preoperative planning, and simulation-based training. 

Accurately predicting implant positions from CBCT scans faces significant technical challenges, primarily stemming from two interconnected aspects of the problem. First, optimal implant positioning heavily depends on the texture and morphological features of adjacent teeth to infer a biomechanically and biologically sound placement. Paradoxically, the target implant site itself is often an edentulous or anatomically ambiguous region, lacking clear local landmarks. This inherent characteristic demands that the predictive model effectively capture long-range dependencies between distant pixels—for instance, reasoning from the roots and crowns of adjacent teeth to the planned implant socket. Conventional Convolutional Neural Networks (CNNs)~\cite{he2016deep,xie2017aggregated}, with their inherently localized receptive fields, struggle to model these essential global contextual relationships. While Vision Transformers (ViTs)~\cite{dosovitskiy2020image} are designed for such long-range interactions via self-attention, their computational complexity scales quadratically with input size, rendering them prohibitively expensive for high-resolution 3D CBCT volumes~\cite{perera2024segformer3d,vaswani2017attention}. 
The second major challenge involves the underutilization of intrinsic geometric constraints in existing data-driven approaches. Specifically, the implant's slope (angulation) is a geometric parameter derived directly from the 3D coordinates of its apex and base. This relationship provides a strong prior that should inherently regularize the position prediction. While some previous works~\cite{yang2023tcslot,yang2026regfreenet} have recognized the importance of slope, they often treat its prediction as a task separate from coordinate regression. This methodological decoupling fails to leverage the intrinsic correlation between the implant's position and its orientation, consequently limiting the model's capacity to learn anatomically consistent configurations.

To address the fundamental limitation in long-range dependency modeling, we introduce the Mamba architecture, which is grounded in State Space Models (SSMs)~\cite{gu2024mamba,liu2024vmamba,kalman1960new}, into our framework. Mamba achieves efficient global context capture through selective scanning mechanisms, avoiding the quadratic computational cost associated with Transformer-based self-attention. 
Capitalizing on this advantage, we design a hybrid encoder, which strategically integrates CNN layers with Mamba modules. Each convolutional block is augmented with a Mamba module. This design enables the network to hierarchically extract local anatomical features via CNNs while simultaneously modeling global contextual dependencies across the scan volume through Mamba's selective scan operations, thereby providing a more comprehensive understanding of the implant site.

To tackle the second challenge of incorporating geometric constraints, we propose a Slope-Coupled Prediction Branch (SCP). This branch explicitly couples the prediction of the implant’s position with its orientation.
Specifically, the SCP generates a position heatmap from the network’s output, identifies the multi-level encoder features that best align with the implant axis, and uses them to predict the slope.
This end-to-end design ensures internal consistency and anatomical plausibility, effectively enforcing a coherent relationship between the predicted implant location and its angulation.

Main contributions of this paper can be summarized as follows:
1) We proposed a hybrid network, ImplantMamba, which significantly enhances the network's perception capabilities without introducing additional computational burden. 
2) A slope-coupled prediction branch is designed to build up the relationship between the implant position and slope, ensuring internal consistency and anatomical plausibility. 
3) Extensive experiments on a large dental implant dataset demonstrated that the proposed ImplantMamba achieves superior performance than the existing methods.

\begin{figure*}
	\centering
	\includegraphics[width=1.0\linewidth]{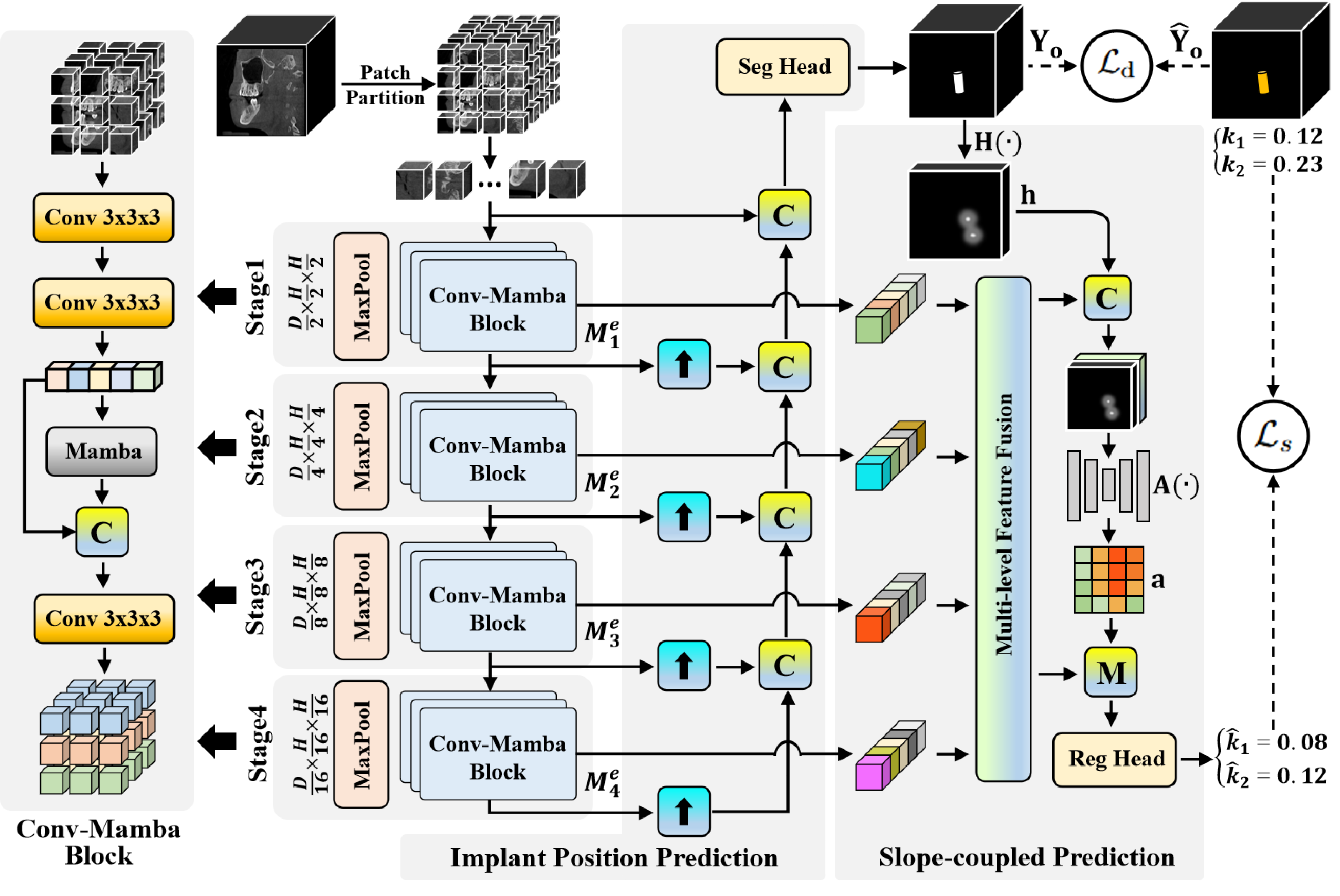}
	\caption{Overview of the proposed ImplantMamba.
    } \label{fig_network}
\end{figure*}

\section{ImplantMamba}
An overview of the proposed ImplantMamba is given in Fig.~\ref{fig_network}. It mainly consists of a hybrid encoder, an implant position prediction branch, and a slope-coupled prediction branch. 
Given a patient's CBCT scan $\mV \in \mathbb{R}^{C\times D\times H\times W}$, the hybrid encoder first learn a series of multi-level fine-grained features $\mM$. 
The implant position prediction branch and the slope-coupled prediction branch take $\mM$ as input to predict the implant position and implant slope, respectively. 
Next, we will introduce these modules in detail.

\subsection{Hybrid Encoder}
Implant location prediction relies on pre-implantation CBCT scans and must be performed in tooth gap areas that lack distinctive texture. This heavy reliance on the texture features of adjacent teeth demands exceptional long-range modeling capabilities from the model~\cite{yang2022implantformer}. 
Although CNNs perform well in local feature extraction, conventional neural networks using uniform convolutions are not effective in perceiving the texture variation between the neighboring teeth.
While Transformer performs well in long-range sequential modeling, its require quadratic cost of self-attention.
Unlike Transformer, Mamba captures global context efficiently via a selective scanning mechanism, which achieves linear computational complexity. 
To leverage the strengths of both architectures, we designed a hybrid encoder comprising multiple Conv-Mamba blocks. Each block sequentially stacks two convolutional layers for local feature extraction followed by one Mamba layer for global context modeling. This design allows the network to hierarchically extract local anatomical features using the CNN layers. Simultaneously, the Mamba layers model global contextual dependencies throughout the scan volume via selective scan operations. The synergistic combination of these components yields a more comprehensive understanding of the implant site.

The architecture of the proposed hybrid encoder is illustrated in Fig.~\ref{fig_network}. It consists of four Conv-Mamba blocks. The encoder progressively downsamples the input features four times, resulting in four output feature maps at different resolutions, denoted as $\{\mM^e_1, \mM^e_2, \mM^e_3, \mM^e_4\}$.

\subsection{Implant Position Prediction Branch}
The implant position prediction branch consists of four upsampling convolutions and a segmentation head.
First, the minimum-resolution encoder feature $\mM^e_4$ is upsampled and then concatenated and fused with the feature $\mM^e_3$.
This operation is repeated four times to restore the feature map resolution to the sub-volume size.
Finally, the prediction head reduces the number of feature channels to 1 to obtain the final implant position. 
Dice loss is used to supervise the implant position prediction branch:
\begin{flalign}
\mathcal{L}_{\text{d}} 
	&=1-\sum\frac{2\sum_{o=1}^{O} \bm{Y}_{o}\cdot\hat{\bm{Y}}_{o}}{\sum_{o=1}^{O} \bm{Y}_{o}^{2}+\sum_{o=1}^{O}\hat{\bm{Y}}_{o}^{2}}, 
\end{flalign}
where $\mO$ represents the number of voxels; $\mY_{\vo}$ and $\hat \mY_{\vo}$ denote the ground truth map and predicted probability map, respectively, at voxel $\vo$.

\subsection{Slope-coupled Prediction Branch}
In clinical practice, the varying widths of alveolar bone among patients necessitate that dentists determine both the position and the angulation (posture) of an implant during surgery. Mathematically, the implant angulation can be represented as a slope derived from its positional coordinates. This implies a close relationship between the implant's slope and its location; the slope serves as a strong prior that inherently regularizes position prediction.
To explicitly leverage this relationship, we design a Slope-Coupled Prediction Branch (SCP) that couples the prediction of the implant’s position with its slope. An overview of the SCP is shown in Fig.~\ref{fig_network}.
Specifically, we first integrate multi-level features
$\{\mM^e_1, \mM^e_2, \mM^e_3, \mM^e_4\}$ from the encoder to obtain a multi-scale feature representation $\mM_{s} \in \mathbb{R}^{C'\times H'\times W'}$. 
A heatmap $\vh \in \mathbb{R}^{H\times W}$, which localizes the implant apex and base, is then generated from the segmentation output to enhance position awareness:
\begin{equation}
\vh = \mH(\hat{\bm{Y}}_{o}),
\end{equation}
where $\mH(\cdot)$ denotes the heatmap generation function. The heatmap $\vh$ is fused with the multi-scale feature $\mM_{s}$. This integration is particularly important in dental CBCT analysis, as optimal implant placement depends on both local bone microstructure (e.g., trabecular density) and global craniofacial anatomy.
Subsequently, an attention network $\mA(\cdot)$ generates a weight matrix to emphasize features most relevant to slope estimation:
\begin{equation}
	\va = \mA(\vh),
\end{equation}
Finally, the attention weights $\va \in \mathbb{R^{H'\times W'}}$ are multiplied with $\mM_{s}$ to selectively highlight features for slope prediction. This design effectively couples implant position with slope, strengthening their interaction. The SCP branch is supervised using an L1 loss:
\begin{equation}
	\mathcal{L}_{s} = |\vk_i - \hat \vk_i|,
\end{equation}
where $\vk_i$ and $\hat \vk_i$ denote the ground-truth and predicted implant slopes, respectively.

\section{Experiments and Results}
\subsection{Dataset and Implementation Details}
We evaluated ImplantMamba on a publicly large-scale CBCT implant dataset, i.e., ImplantFairy~\cite{yang2026regfreenet}.
It contains 1622 CBCT datasets, covering individuals from youth, middle-aged, and elderly groups. We followed the official partitioning to divide the training and test sets, with 1369 datasets in the training set and 253 in the test set.
All models were implemented using the MONAI framework, with an input size of 128×128×128. The learning rate was set to 1×$10^{-4}$ and adjusted via a cosine warmup schedule.
Given the anatomical specificity of the oral cavity, data augmentation was limited to random cropping.

\subsection{Performance Analysis}

\begin{figure*}
	\centering
	\includegraphics[width=0.9\linewidth]{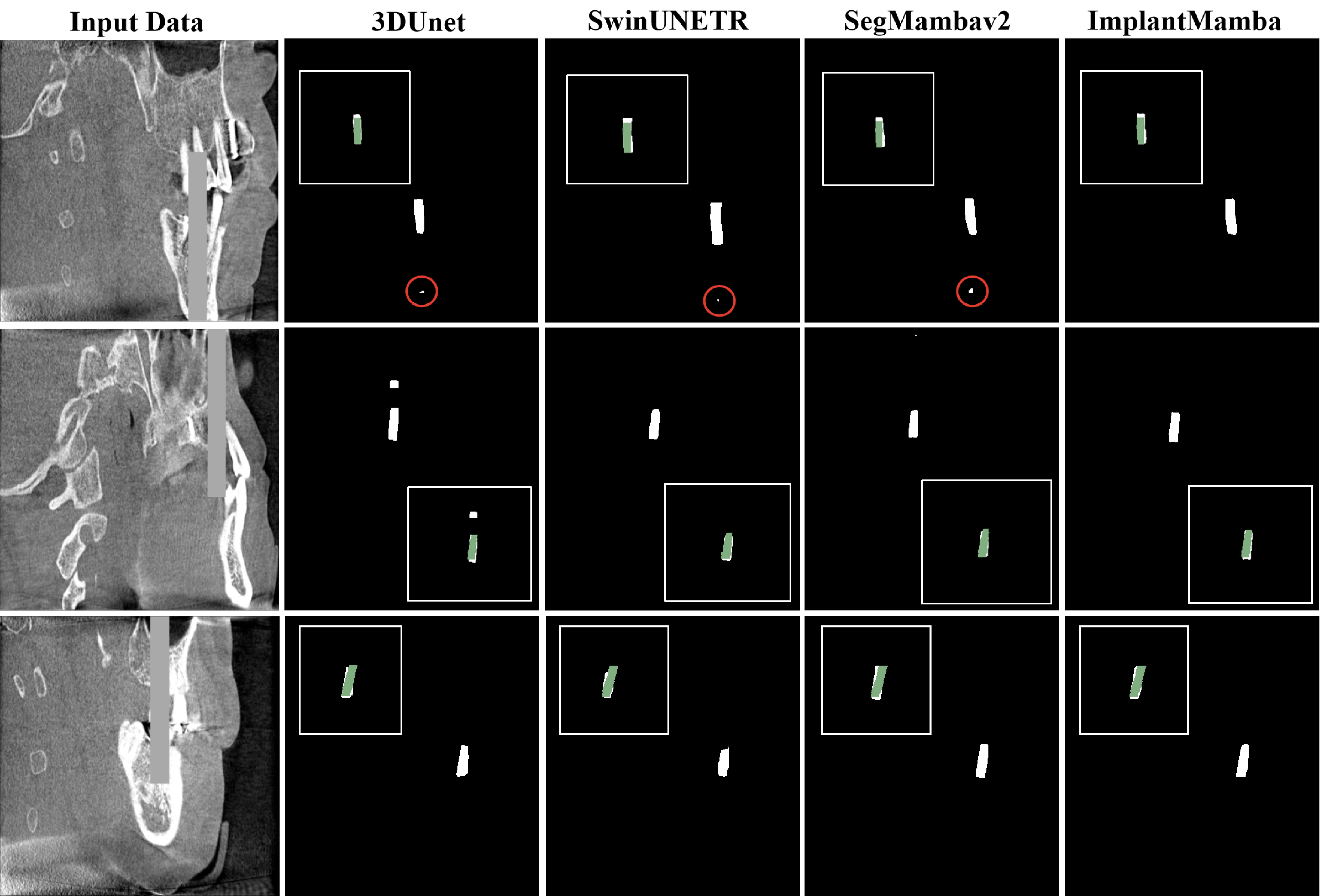}
	\caption{Visualization of the predicted implant on the ImplantFairy dataset. The white and green masks represent the predicted implant and the actual implant, respectively.} \label{fig_vis}
\end{figure*}

\subsubsection{Ablation Studies.} 
To evaluate the impact of integrating the proposed Conv-Mamba block into the hybrid encoder, we conducted an ablation study by progressively integrating them across the encoder's layers, as detailed in Table~\ref{tab_ablation}.
The results reveal a notable and counter-intuitive trend: incorporating the Conv-Mamba block into more layers leads to a degradation in segmentation performance. The network achieves its optimal performance—a Dice score of 48.41\% and an IoU of 0.359—when the Conv-Mamba block is used only in the first layer. Performance declines as modules are added to subsequent layers, reaching the lowest scores (Dice: 45.49\%) when blocks are included in the first two layers.
This clear downward trend indicates that while the Mamba module benefits the initial feature extraction stage, its incorporation into deeper layers is detrimental. 

Furthermore, we verified the effectiveness of the SCP branch. As shown in Table 1, applying SCP on top of the hybrid encoder further improves performance, with the optimal result achieved when using only a single Conv-Mamba block alongside SCP (Dice: 48.41\%, IoU: 0.359). The experimental results demonstrate that the SCP branch can assist the hybrid encoder in extracting more effective features.


\begin{table}[htbp]
\centering
\caption{Comparison to the mainstream segmentation methods}
\label{tab_ablation}
\begin{tabular}{cccccc cc}
\toprule
\multicolumn{4}{c}{\textbf{Conv-Mamba Block}} & \textbf{SCP} & \textbf{Dice(\%)} & \textbf{IoU} \\
\cmidrule(lr){1-4}
Layer 1 & Layer 2 & Layer 3 & Layer 4 & & & \\
\midrule
$\times$& $\times$&$\times$ &$\times$ &$\times$ & 45.48 & 0.336 \\
$\checkmark$ &$\times$ &$\times$&$\times$ &$\times$ & 46.83 & 0.344 \\
$\checkmark$ & $\checkmark$ &$\times$ &$\times$ &$\times$ & 45.56 & 0.337 \\
$\checkmark$ & $\checkmark$ & $\checkmark$ &$\times$ & $\times$& 45.73 & 0.337 \\
$\checkmark$ & $\checkmark$ & $\checkmark$ & $\checkmark$  & $\times$& 45.90 & 0.338 \\
$\checkmark$ & $\times$& $\times$& $\times$& $\checkmark$ & \textbf{48.41} & \textbf{0.359} \\
$\checkmark$ & $\checkmark$ & $\times$& $\times$&$\checkmark$ & 47.49 & 0.354 \\
$\checkmark$ & $\checkmark$ & $\checkmark$ & $\times$& $\checkmark$ & 47.14 & 0.350 \\
$\checkmark$ & $\checkmark$ & $\checkmark$ & $\checkmark$  & $\checkmark$ & 45.49 & 0.336 \\
\bottomrule
\end{tabular}
\end{table}

\subsubsection{Comparison to the State-of-the-art Methods.}
To verify the effectiveness of our method, we compare ImplantMamba against the state-of-the-art segmentation methods, covering CNN-based, Transformer-based, and Mamba-based architectures. 
The compared CNN-based methods include 3DUNet~\cite{ronneberger2015u}, 3DUNet++~\cite{zhou2018unet++}, DAF3D~\cite{wang2019deep}, VNet~\cite{milletari2016v} and UXNet~\cite{lee20223d}.
Transformer-based methods include UNETR~\cite{hatamizadeh2022unetr}, UNETR++~\cite{shaker2024unetr++} and SwinUNETR~\cite{hatamizadeh2021swin}. 
Mamba-based methods include SegMamba~\cite{xing2024segmamba}, SegMamba v2~\cite{xing2025segmamba} and EMNet~\cite{chang2024net}.
The transformer-based methods utilize Vision Transformer~\cite{dosovitskiy2020image} or SwinTransformer~\cite{liu2021swin} as encoders to learn global features for prediction.
To ensure fairness, we use publicly available codes of all these methods and train them under the same training settings.
Experimental results are given in Table~\ref{tab_compare}. 
The evaluation metrics include model parameter count (Param), Dice coefficient (Dice), and Intersection over Union (IoU).

\begin{table}[htbp]
  \centering
  \caption{Comparison to the mainstream segmentation methods.}
  \label{tab_compare}
  \begin{tabular}{ccccc}
    \toprule
    \textbf{Method} & \textbf{Model} & \textbf{Param(M)} & \textbf{Dice(\%)} & \textbf{IoU} \\
    \midrule
    \multirow{5}{*}{\textbf{CNN-based}}
      & 3DUNet~\cite{ronneberger2015u}    & \textbf{5.7}   & 45.48 & 0.336 \\
      & 3DUNet++~\cite{zhou2018unet++}  & 6.9   & 43.75 & 0.316 \\
      & DAF3D~\cite{wang2019deep}     & 29.2  & 44.88 & 0.319 \\
      & VNet~\cite{milletari2016v}      & 45.6  & 43.23 & 0.311 \\
      & UXNet~\cite{lee20223d}     & 53.0  & 42.09 & 0.309 \\
    \midrule
    \multirow{3}{*}{\textbf{Transformer-based}}
      & UNETR~\cite{hatamizadeh2022unetr}     & 93.0  & 41.34 & 0.282 \\
      & UNETR++~\cite{shaker2024unetr++}   & 122.1 & 43.8  & 0.316 \\
      & SwinUNETR~\cite{hatamizadeh2021swin} & 62.1  & 44.03 & 0.321 \\
    \midrule
    \multirow{4}{*}{\textbf{Mamba-based}}
      & SegMamba~\cite{xing2024segmamba}        & 67.4  & 46.39 & 0.341 \\
      & SegMambav2~\cite{xing2025segmamba}      & 138.7 & 46.56 & 0.345 \\
      & EMNet~\cite{chang2024net}          & 61.5  & 44.26 & 0.322 \\
      & ImplantMamba(Ours) & 7.8 & \textbf{48.41} & \textbf{0.359} \\
    \bottomrule
  \end{tabular}
\end{table}

From the table we can observe several key trends. First, among CNN-based models, approaches like 3DUNet achieve moderate Dice scores (e.g., 45.48\%) with relatively low parameter counts (5.7M). However, as architectural complexity increases—exemplified by UXNet (53.0M)—performance does not consistently improve and can even decline (Dice of 42.09\%). This suggests potential limitations in capturing long-range dependencies, which are critical for this task.
Second, Transformer-based models, such as UNETR and SwinUNETR, incorporate significant parameter overhead (93.0M and 62.1M parameters, respectively) to model global context. While they outperform certain CNNs, their Dice scores remain suboptimal (ranging from 41.34\% to 44.03\%). Notably, SegMamba, a recent hybrid approach, achieves better performance (46.39\% Dice) with relatively fewer parameters (67.4M) than most pure Transformer models.
Importantly, our ImplantMamba achieves the best performance on both primary metrics, with a Dice score of 48.41\% and an IoU of 0.359, while maintaining high parameter efficiency (7.8M). 
ImplantMamba accomplishes this with only a fraction of the parameters required by most Transformer and advanced Mamba models, highlighting its superior design in balancing model capacity, efficiency, and ability to capture long-range spatial dependencies.

\subsubsection{Visualization.}
To further validate the performance of ImplantMamba, we selected the best-performing models from the CNN-based (3DUnet), Transformer-based (SwinUNETR), and Mamba-based (SegMambav2) categories for a visual comparison of prediction results, as shown in Fig.~\ref{fig_vis}.
The first row visualizes the implant prediction for the right mandibular jaw. It can be observed that all compared methods, except for ImplantMamba, exhibit areas of false positives (marked by red circles). Furthermore, in terms of predicting implant slope, ImplantMamba's result aligns more closely with the ground truth. The second row presents the results for the left maxillary jaw. Here, only 3DUnet shows clear false positives, which highlights the limited contextual modeling capability of pure CNN-based architectures. Additionally, both SwinUNETR and SegMambav2 demonstrate inferior performance compared to ImplantMamba in localizing the implant's start and end points. The third row displays a challenging case involving a highly tilted implant in the right maxillary jaw. While SwinUNETR correctly captures the implant's general posture, it mispredicts its precise start and end locations. Conversely, SegMambav2 achieves better accuracy in endpoint localization, yet ImplantMamba provides a slope prediction that is closer to the true value.
These experimental results demonstrate that our proposed ImplantMamba can effectively utilize the inclination constraint to refine implant position prediction, thereby contributing to its superior overall performance.

\section{Conclusion}
In this study, we present a novel implant position prediction network (ImplantMamba), which integrates long-range contextual modeling with slope-aware regression. 
To address the challenge of texture-deficient implant regions in CBCT images, our approach incorporates a hybrid CNN-Mamba encoder to hierarchically capture both local anatomical features and global dependencies across the scan volume. 
A dedicated Slope-Coupled Prediction (SCP) branch is introduced to enforce anatomical consistency between the implant position and its slope. 
Extensive experiments on a large-scale dental implant dataset confirm that ImplantMamba achieves state-of-the-art performance, demonstrating its effectiveness in supporting surgical guide design.

\bibliographystyle{splncs04}
\bibliography{ref}

@inproceedings{yang2023tcslot,
  title={Tcslot: Text guided 3d context and slope aware triple network for dental implant position prediction},
  author={Yang, Xinquan and Xie, Jinheng and Li, Xuechen and Li, Xuguang and Shen, Linlin and Deng, Yongqiang},
  booktitle={2023 IEEE International Conference on Bioinformatics and Biomedicine (BIBM)},
  pages={726--732},
  year={2023},
  organization={IEEE}
}

@article{nazir2020global,
  title={Global prevalence of periodontal disease and lack of its surveillance},
  author={Nazir, Muhammad and Al-Ansari, Asim and Al-Khalifa, Khalifa and Alhareky, Muhanad and Gaffar, Balgis and Almas, Khalid},
  journal={The Scientific World Journal},
  volume={2020},
  year={2020},
  publisher={Hindawi}
}

@inproceedings{milletari2016v,
	title={V-net: Fully convolutional neural networks for volumetric medical image segmentation},
	author={Milletari, Fausto and Navab, Nassir and Ahmadi, Seyed-Ahmad},
	booktitle={2016 fourth international conference on 3D vision (3DV)},
	pages={565--571},
	year={2016},
	organization={Ieee}
}

@article{kernen2020review,
  title={A review of virtual planning software for guided implant surgery-data import and visualization, drill guide design and manufacturing},
  author={Kernen, Florian and Kramer, Jaap and Wanner, Laura and Wismeijer, Daniel and Nelson, Katja and Fl{\"u}gge, Tabea},
  journal={BMC Oral health},
  volume={20},
  number={1},
  pages={251},
  year={2020},
  publisher={Springer}
}

@inproceedings{xie2017aggregated,
  title={Aggregated residual transformations for deep neural networks},
  author={Xie, Saining and Girshick, Ross and Doll{\'a}r, Piotr and Tu, Zhuowen and He, Kaiming},
  booktitle={Proceedings of the IEEE conference on computer vision and pattern recognition},
  pages={1492--1500},
  year={2017}
}

@article{vaswani2017attention,
  title={Attention is all you need},
  author={Vaswani, Ashish and Shazeer, Noam and Parmar, Niki and Uszkoreit, Jakob and Jones, Llion and Gomez, Aidan N and Kaiser, {\L}ukasz and Polosukhin, Illia},
  journal={Advances in neural information processing systems},
  volume={30},
  year={2017}
}

@inproceedings{perera2024segformer3d,
  title={Segformer3d: an efficient transformer for 3d medical image segmentation},
  author={Perera, Shehan and Navard, Pouyan and Yilmaz, Alper},
  booktitle={Proceedings of the IEEE/CVF Conference on Computer Vision and Pattern Recognition},
  pages={4981--4988},
  year={2024}
}

@article{kalman1960new,
  title={A new approach to linear filtering and prediction problems},
  author={Kalman, Rudolph Emil},
  year={1960}
}

@inproceedings{gu2024mamba,
  title={Mamba: Linear-time sequence modeling with selective state spaces},
  author={Gu, Albert and Dao, Tri},
  booktitle={First conference on language modeling},
  year={2024}
}

@article{liu2024vmamba,
  title={Vmamba: Visual state space model},
  author={Liu, Yue and Tian, Yunjie and Zhao, Yuzhong and Yu, Hongtian and Xie, Lingxi and Wang, Yaowei and Ye, Qixiang and Jiao, Jianbin and Liu, Yunfan},
  journal={Advances in neural information processing systems},
  volume={37},
  pages={103031--103063},
  year={2024}
}

@article{wang2019deep,
	title={Deep attentive features for prostate segmentation in 3D transrectal ultrasound},
	author={Wang, Yi and Dou, Haoran and Hu, Xiaowei and Zhu, Lei and Yang, Xin and Xu, Ming and Qin, Jing and Heng, Pheng-Ann and Wang, Tianfu and Ni, Dong},
	journal={IEEE transactions on medical imaging},
	volume={38},
	number={12},
	pages={2768--2778},
	year={2019},
	publisher={IEEE}
}

@inproceedings{zhou2018unet++,
	title={Unet++: A nested u-net architecture for medical image segmentation},
	author={Zhou, Zongwei and Rahman Siddiquee, Md Mahfuzur and Tajbakhsh, Nima and Liang, Jianming},
	booktitle={International workshop on deep learning in medical image analysis},
	pages={3--11},
	year={2018},
	organization={Springer}
}

@inproceedings{ronneberger2015u,
	title={U-net: Convolutional networks for biomedical image segmentation},
	author={Ronneberger, Olaf and Fischer, Philipp and Brox, Thomas},
	booktitle={International Conference on Medical image computing and computer-assisted intervention},
	pages={234--241},
	year={2015},
	organization={Springer}
}

@article{shaker2024unetr++,
	title={UNETR++: delving into efficient and accurate 3D medical image segmentation},
	author={Shaker, Abdelrahman and Maaz, Muhammad and Rasheed, Hanoona and Khan, Salman and Yang, Ming-Hsuan and Khan, Fahad Shahbaz},
	journal={IEEE Transactions on Medical Imaging},
	volume={43},
	number={9},
	pages={3377--3390},
	year={2024},
	publisher={IEEE}
}

@inproceedings{hatamizadeh2021swin,
	title={Swin unetr: Swin transformers for semantic segmentation of brain tumors in mri images},
	author={Hatamizadeh, Ali and Nath, Vishwesh and Tang, Yucheng and Yang, Dong and Roth, Holger R and Xu, Daguang},
	booktitle={International MICCAI brainlesion workshop},
	pages={272--284},
	year={2021},
	organization={Springer}
}

@inproceedings{hatamizadeh2022unetr,
	title={Unetr: Transformers for 3d medical image segmentation},
	author={Hatamizadeh, Ali and Tang, Yucheng and Nath, Vishwesh and Yang, Dong and Myronenko, Andriy and Landman, Bennett and Roth, Holger R and Xu, Daguang},
	booktitle={Proceedings of the IEEE/CVF winter conference on applications of computer vision},
	pages={574--584},
	year={2022}
}

@article{lee20223d,
	title={3d ux-net: A large kernel volumetric convnet modernizing hierarchical transformer for medical image segmentation},
	author={Lee, Ho Hin and Bao, Shunxing and Huo, Yuankai and Landman, Bennett A},
	journal={arXiv preprint arXiv:2209.15076},
	year={2022}
}

@article{liu2021transfer,
  title={Transfer Learning via Artificial Intelligence for Guiding Implant Placement in the Posterior Mandible: an in vitro Study},
  author={Liu, Yun and Chen, Zhi-cong and Chu, Chun-ho and Deng, Fei-Long},
  year={2021}
}

@article{dosovitskiy2020image,
  title={An image is worth 16x16 words: Transformers for image recognition at scale},
  author={Dosovitskiy, Alexey and Beyer, Lucas and Kolesnikov, Alexander and Weissenborn, Dirk and Zhai, Xiaohua and Unterthiner, Thomas and Dehghani, Mostafa and Minderer, Matthias and Heigold, Georg and Gelly, Sylvain and others},
  journal={arXiv preprint arXiv:2010.11929},
  year={2020}
}

@article{yang2023two,
  title={Two-Stream Regression Network for Dental Implant Position Prediction},
  author={Yang, Xinquan and Li, Xuguang and Li, Xuechen and Chen, Wenting and Shen, Linlin and Li, Xin and Deng, Yongqiang},
  journal={Expert Systems with Applications},
  year={2023}
}

@article{elani2018trends,
  title={Trends in dental implant use in the US, 1999--2016, and projections to 2026},
  author={Elani, HW and Starr, JR and Da Silva, JD and Gallucci, GO},
  journal={Journal of dental research},
  volume={97},
  number={13},
  pages={1424--1430},
  year={2018},
  publisher={SAGE Publications Sage CA: Los Angeles, CA}
}

@article{yang2022implantformer,
  title={ImplantFormer: Vision Transformer based Implant Position Regression Using Dental CBCT Data},
  author={Yang, Xinquan and Li, Xuguang and Li, Xuechen and Wu, Peixi and Shen, Linlin and Li, Xin and Deng, Yongqiang},
  journal={arXiv preprint arXiv:2210.16467},
  year={2022}
}

@inproceedings{he2016deep,
  title={Deep residual learning for image recognition},
  author={He, Kaiming and Zhang, Xiangyu and Ren, Shaoqing and Sun, Jian},
  booktitle={Proceedings of the IEEE conference on computer vision and pattern recognition},
  pages={770--778},
  year={2016}
}

@article{yang2026regfreenet,
  title={RegFreeNet: A Registration-Free Network for CBCT-based 3D Dental Implant Planning},
  author={Yang, Xinquan and Li, Xuguang and Zheng, Mianjie and Liu, Xuefen and Tang, Kun and Lim, Kian Ming and Meng, He and Ren, Jianfeng and Shen, Linlin},
  journal={arXiv preprint arXiv:2601.14703},
  year={2026}
}

@inproceedings{xing2024segmamba,
  title={SegMamba: Long-Range Sequential Modeling Mamba for 3D Medical Image Segmentation},
  author={Xing, Zhaohu and Ye, Tian and Yang, Yijun and Liu, Guang and Zhu, Lei},
  booktitle={International Conference on Medical Image Computing and Computer-Assisted Intervention},
  pages={578--588},
  year={2024}
}

@article{xing2025segmamba,
  title={Segmamba-v2: Long-range sequential modeling mamba for general 3d medical image segmentation},
  author={Xing, Zhaohu and Ye, Tian and Yang, Yijun and Cai, Du and Gai, Baowen and Wu, Xiao-Jian and Gao, Feng and Zhu, Lei},
  journal={IEEE Transactions on Medical Imaging},
  year={2025},
  publisher={IEEE}
}

@inproceedings{chang2024net,
  title={Em-net: Efficient channel and frequency learning with mamba for 3d medical image segmentation},
  author={Chang, Ao and Zeng, Jiajun and Huang, Ruobing and Ni, Dong},
  booktitle={International Conference on Medical Image Computing and Computer-Assisted Intervention},
  pages={266--275},
  year={2024},
  organization={Springer}
}

@inproceedings{liu2021swin,
  title={Swin transformer: Hierarchical vision transformer using shifted windows},
  author={Liu, Ze and Lin, Yutong and Cao, Yue and Hu, Han and Wei, Yixuan and Zhang, Zheng and Lin, Stephen and Guo, Baining},
  booktitle={Proceedings of the IEEE/CVF international conference on computer vision},
  pages={10012--10022},
  year={2021}
}

\end{document}